\documentclass[a4paper,twoside]{article}

\usepackage{epsfig}
\usepackage{subcaption}
\usepackage{calc}
\usepackage{amssymb}
\usepackage{amstext}
\usepackage{amsmath}
\usepackage{amsthm}
\usepackage{multicol}
\usepackage{pslatex}
\usepackage{apalike}
\usepackage{algorithm2e}
\usepackage[bottom]{footmisc}
\usepackage{SCITEPRESS}     
\usepackage{tabularx}
\usepackage[table]{xcolor}

\begin{document}

\title{Enactor: From Traffic Simulators to Surrogate World Models}

\author{\authorname{Yash Ranjan\sup{1}, Rahul Sengupta\sup{1}, Anand Rangarajan\sup{1} and Sanjay Ranka\sup{1}}
\affiliation{\sup{1}Department of CISE, University of Florida}
\email{\{yashranjan, rahulseng\}@ufl.edu, \{anand, ranka\}@cise.ufl.edu}
}


\abstract{Traffic microsimulators such as SUMO are widely used to evaluate road network performance under various ``what-if” conditions. However, the behavior models controlling the actions of the actors are overly simplistic and fails to capture realistic actor-actor interactions and its impact on an actors behavior. Deep learning-based methods have been applied to model vehicles and pedestrians as ``agents'' responding to their surrounding ``environment'' (including lanes, signals, and neighboring agents). Although effective in learning actor-actor interaction, these approaches fail to generate physically consistent trajectories over long time periods, and they do not explicitly address the complex dynamics that arise at traffic intersections which is a critical location in urban networks. Inspired by the World Model paradigm, we have developed an actor centric generative model using transformer-based architecture that is able to capture the actor-actor interaction, at the same time understanding the geometry to the traffic intersection to generate physically grounded trajectories that are based on learned behavior from the data. Moreover, we test the model in a live ``simulation-in-the-loop'' setting, where we generate the initial conditions of the actors using SUMO and then let the model control the dynamics of the actors. We let the simulation run for 40000 timesteps (4000 seconds), testing the performance of the model on long timerange and evaluating the trajectories on traffic engineering related metrics. Experimental results demonstrate that the proposed framework effectively captures complex actor--actor interactions and generates long-horizon, physically consistent trajectories, while requiring significantly fewer training samples than traditional agent-centric generative approaches. Our model is able to outperform the baseline in most of the metrics as well as aggregate speed and travel-time metrics where our model beats the baseline by more than 10x on the KL-Divergence.}

\onecolumn \maketitle \normalsize \setcounter{footnote}{0} \vfill

\section{\uppercase{Introduction}}
\label{sec:introduction}

Microsimulators for transportation are used to model large geographical regions by constructing a virtual representation of the environment and simulating the behavior of individual actors within it. By explicitly modeling each agent, microsimulation enables the study of how changes to infrastructure, control strategies, or demand patterns influence a wide range of system-level traffic statistics. These simulators provide a testbed for conducting experiments in silico, allowing the impact of potential interventions to be evaluated before they are deployed in the real world, thereby reducing the risk of undesirable consequences and inefficiencies. There are several widely used open-source platforms such as SUMO and MATSim which offers researchers accessible tools to analyze and study traffic flow at scale. 
Using microsimulators, researchers can investigate a wide range of transportation-related questions, such as evaluating alternative intersection and corridor designs including lane additions, roundabouts, turn lanes, and signal phasing schemes and compare their impacts on delays, queue lengths, and travel times. Microsimulation is also commonly used to study driver behavior, car-following and lane-changing dynamics, and safety-related surrogate metrics (e.g., conflicts and near-miss events), particularly for vulnerable road users.

In traffic microsimulation, the state of each actor is updated at every small time step, typically ranging from 0.1 to 1 second using behavior models that react to the actor’s local environment. System-level traffic phenomena such as queues, shockwaves, and congestion emerge from the aggregation of these thousands of localized, individual updates. For vehicle dynamics, commonly used car-following models such as the Intelligent Driver Model and the Krauss model compute a vehicle’s acceleration based on its own speed, the distance to the leading vehicle, and the leader’s velocity. Pedestrian microsimulation, in contrast, often relies on force-based formulations such as the Social Force Model, in which each agent experiences attractive forces toward its goal and repulsive forces from other pedestrians and obstacles.

Despite their effectiveness, these behavior models have notable limitations in capturing realistic human behavior\cite{NI2020102137}. Many car-following and lane-changing models assume “average” drivers with smooth and rational responses, making it difficult to reproduce aggressive, distracted, or abnormal driving behaviors, as well as rare and safety-critical maneuvers. Moreover, calibrating such models is challenging due to limited availability of high-fidelity behavioral data, which restricts the estimation of fine-grained, actor-specific parameters. As a result, these models typically rely on population-level average parameters, limiting their ability to accurately capture diverse real-world actor–actor interactions.

The advent of machine learning, and deep neural networks in particular, has fundamentally transformed modeling paradigms in transportation systems. Data-driven approaches enable the learning of complex, nonlinear dependencies and high-dimensional interaction patterns that are difficult, if not infeasible to capture using hand-crafted, equation-based models. Moreover, probabilistic and generative learning frameworks provide a natural mechanism to represent the inherent uncertainty and stochasticity present in human decision making. As a result, generative architectures have been widely adopted for trajectory forecasting\cite{salzmann2021}\cite{shi2023mtr} and, more recently, for simulating the motion of surrounding vehicles in autonomous driving contexts\cite{trafficbots2023}.

However, simulation frameworks based on purely agent-centric generation face two key challenges. First, closed-loop rollout over long horizons is highly susceptible to error accumulation and covariate shift, as small prediction errors compound over time and progressively shift the system away from the training distribution. Second, learning physically plausible and socially consistent trajectories even over short horizons typically requires large volumes of high-quality data, due to the need to implicitly learn complex physical constraints and multi-agent dynamics from observations alone.

World models\cite{ha2020} constitute a line of research rooted in insights from neuroscience, which suggest that humans internally simulate future states of their environment before committing to an action. By mentally rolling out possible futures, humans can evaluate and discard unfavorable actions without executing them in the real world. This perspective extends beyond the classical Markov Decision Process (MDP) formulation, which typically focuses on directly learning a policy or value function that maps states to actions in order to maximize long-term reward, without explicitly modeling environment dynamics.

In contrast, world-model-based approaches aim to learn an internal, predictive model of the environment itself, enabling reasoning, planning, and decision making through simulated rollouts. Such models have recently gained traction in physical AI settings, including autonomous driving and robotics, where capturing complex environment dynamics and long-horizon consequences is essential for effective planning and control.

Traffic intersections represent highly complex conflict zones where the trajectories of vehicles and pedestrians intersect, and a substantial fraction of both fatal and non-fatal crashes occur. Beyond safety, intersections also exert a significant influence on overall travel times and congestion levels across traffic corridors. Accurately modeling the diverse and often uncertain behaviors of traffic participants at intersections is therefore critical for improving safety, operational efficiency, and driving comfort.

Traffic intersections pose a uniquely challenging simulation setting: they exhibit high uncertainty due to stochastic decision making and rich actor–actor interactions, while simultaneously providing strong structural constraints in the form of fixed geometry, lane topology, and signal timing plans. In this work, we leverage this structure by combining principles from world-model-based learning with generative, multi-agent interaction modeling to develop an intelligent, data-driven microsimulation framework for traffic intersections.

\section{\uppercase{Related Works}}

Significant research has been done on Deep Learning-based Trajectory Prediction Models \cite{trajsurvey1}. These models often use the vehicle's prior trajectory trace, along with information such as neighboring agents, lane geometry etc. Trajectron++~\cite{salzmann2021} was an important model that combined LSTM-based history encoding with graph attention for interaction modeling, but it was not able to capture the dynamics of a traffic intersection such as traffic signal. IntersectionNet~\cite{intersectionnet} similarly unrolls trajectories, incorporating curvilinear coordinates and geometry-aware attention, demonstrating its efficacy at signalized intersections. \cite{hdgt2023} introduces a graph transformer that encodes scene context and agent dynamics for multi-agent trajectory prediction. These models are primarily designed to simply unroll the trajectories under certain conditions.

On the other hand, `World Models' that try to unroll the dynamics of the entire state of the environment have also been gaining popularity in the past few years. Survey papers such as \cite{guan2024worldmodelsautonomousdriving} provide comprehensive surveys and taxonomies of world-model approaches in autonomous driving, covering perception, prediction, planning, and simulation applications.

\cite{trafficbots2023} is a major early work in this field, and it proposes a world model that jointly predicts multi-agent trajectories and scene evolution for autonomous driving. This work presents `TrafficBots' for multi-agent motion prediction and end-to-end driving. Each agent has configurable behaviors, based on destination (navigational information), and a time-invariant latent personality for behavioral. The authors evaluate their model on the Waymo Open Motion dataset \cite{ettinger2021large}. 

\cite{wang2023buildingtransportationfoundationmodel} proposes a foundation model for transportation systems using a generative graph transformer that learns traffic dynamics from large-scale graph-structured data. The authors use this in their work `TransWorldNG' \cite{transworldng2023}, which describes a foundation-model-based traffic simulator that learns traffic dynamics from real-world data using graph learning.

In \cite{tan2025scenediffusercityscaletrafficsimulation}, a city-scale traffic simulation `SceneDiffuser++' is introduced. It uses a Generative World Model, which takes as input the map of the city and an autonomous vehicle (AV) software stack. It then simulates a trip across the city, populating and controlling the scene around the AV, including dynamic agents (vehicles and pedestrians), as well as controlling the traffic lights. The authors assess simulation quality on an augmented version of the Waymo Open Motion Dataset.

Evaluating such models properly is also important.\cite{metrics} develops an intersection-specific evaluation framework, demonstrating that low trajectory reconstruction error does not guarantee rule-compliant behavior (e.g., red-light violations, unsafe time-to-collision events). \cite{beyondsimulation2025} details benchmarking of various world models for autonomous driving and evaluating their robustness as traffic simulators. They  propose new metrics to highlight the sensitivity of world models to uncontrollable vehicles i.e. agents that the model can not control, but still must contend with.

Most of the research in this domain have addressed the problem from an Autonomous Vehicle (AV) perspective, where the aim is to better model the behavior of actors around an AV. Although the architectural innovations have been impressive, most works are still not able to generate long range, physically valid trajectories of a dynamic number of agents. Moreover, none of the research specifically aims to study the complex interactions that occur at a traffic intersection, which is one of the most critical parts of the road transportation infrastructure.

The key contributions of this paper are threefold: (i) a physically constrained, polar coordinate representation that respects roadway geometry, and (ii) a transformer-based interaction modeling and trajectory unrolling architecture trained in closed loop that helps the model generalize better for long range prediction tasks (iii) a set of traffic engineering related metrics that helps measure the performance of the model at multiple traffic intersections.

In the next section, we will define the problem and the proposed architecture. We will then present experiments, followed by their results. We end the paper with the conclusion and future directions of work.

\section{\uppercase{Problem Definition}}
We characterize an intersection using both the actors present within it and a set of control parameters that define the preconditions of the traffic environment. These control parameters are essential because driver and pedestrian behaviors vary significantly across different intersection configurations; thus, defining a structured parameter set enables systematic comparison and modeling across diverse scenarios. In this work, we train our model on two different \textbf{intersection geometry} to verify if our model is able to learn driving behavior across multiple intersection. In future work we would also like to see how \textbf{inbound traffic rate or density} which captures the flow characteristics entering the intersection effect driving behavior.

We can further characterize an intersection by the set of actors present within the scene. These include \emph{static actors}, such as lane centerlines, lane boundaries, and other infrastructural elements, as well as \emph{dynamic actors}, namely vehicles, pedestrians, and the traffic signal phases. In this work, we adopt an actor-centric modeling perspective for vehicles, wherein the future states of each agent are predicted based on its interactions with surrounding actors.

Following \cite{montali2023}, we formulate driving as a Hidden Markov Model (HMM) 
\[
\mathcal{H} = \langle \mathcal{S}, \mathcal{O}, p(o_t \mid s_t), p(s_t \mid s_{t-1}) \rangle,
\]
where $\mathcal{S}$ denotes the set of unobservable true world states, $\mathcal{O}$ denotes the set of observations, $p(o_t \mid s_t)$ is the sampleable emission distribution, and $p(s_t \mid s_{t-1})$ represents the hidden Markovian state dynamics governing the transition from state $s_{t-1}$ to $s_t$.

We denote the true observation dynamics as
\[
p_{\mathrm{world}}(o_t \mid s_{t-1}) =
\mathbb{E}_{p(s_t \mid s_{t-1})\, p(o_t \mid s_t)},
\]
which captures the generative process underlying the observed scene evolution. The goal is to learn a \emph{world model} 
\[
q_{\mathrm{world}}(o_t \mid o^{c}_{<t}),
\]
that approximates $p_{\mathrm{world}}(o_t \mid s_{t-1})$, where the context
\[
o^{c}_{<t} = [\text{omap},\, \text{osignals},\, o_{-H-1}, \ldots, o_{t-1}]
\]
consists of the static map, traffic signal observations, and a history of past observations of length $H$.

For an agent-centric formulation, we adapt the world-model definition by modeling the conditional distribution
\begin{equation}
q_{\mathrm{agent}}(o^i_t \mid o^{c(i)}_{<t}),
\end{equation}
where the context for agent $i$ at time $t$ is defined as 
\begin{equation}
o^{c(i)}_t = [\, map^i_t,\, signal^i_t,\, o^i_t,\, n^{1..N(i,t)}_t \,].
\end{equation}
Here, $o^i_t$ denotes the observed state of actor $i$ at time $t$, and $n^j_t$ represents the state of its $j$-th neighbor at time $t$, with $N(i,t)$ denoting the total number of neighbors for actor $i$. The terms $map^i_t$ and $signal^i_t$ correspond to the local map features and traffic signal information in the vicinity of actor $i$.

Following \cite{zhang2025}, we adopt a polar-coordinate-based representation for dynamic actors to better capture the curved trajectories commonly observed in traffic intersections. The state of actor $i$ and its neighbors is represented as
\begin{equation}
o^i_t = (r,\, \sin\theta,\, \cos\theta,\, s,\, \sin\alpha,\, \cos\alpha),
\end{equation}
where $r$ is the radial distance from the intersection center, $\theta$ is the angular position, $s$ is the linear speed, and $\alpha$ is the heading angle of the velocity vector.

For static map elements, we adopt the polyline-based HD-map representation of \cite{gao2020}, which encodes roadway geometry as sets of vectorized lane-centerline or boundary segments. The local map for actor $i$ is represented as
\begin{equation}
map^i_t \in \mathbb{R}^{P^i_t \times L \times 6},
\end{equation}
where $P^i_t$ is the number of polylines in the vicinity of actor $i$ and $L$ is the number of vectors per polyline. Each vector is encoded as
\begin{equation}
(r,\, \sin\theta,\, \cos\theta,\, \ell,\, \sin\alpha,\, \cos\alpha),
\end{equation}
with $\ell$ denoting the segment length and $\alpha$ the orientation of the vector. This representation provides a compact yet expressive parameterization of local roadway geometry while maintaining consistency with the actor-state polar coordinate system.
The traffic signal is represented by the lane end vector with the one-hot encoded traffic signal state concatenated to it. The traffic signal vector is encoded as
\begin{equation}
(r,\ \sin\theta,\ \cos\theta,\ \ell,\ \sin\alpha,\ \cos\alpha,\ \mathbf{e}_{\text{sig}}),
\end{equation}
where $\mathbf{e}_{\text{sig}}\in\{0,1\}^{K}$ is a one-hot vector indicating the current signal state (e.g., $K=4$ for red/yellow/flashing-yellow/green).
\begin{figure*}[t]
    \centering
    \includegraphics[width=\textwidth]{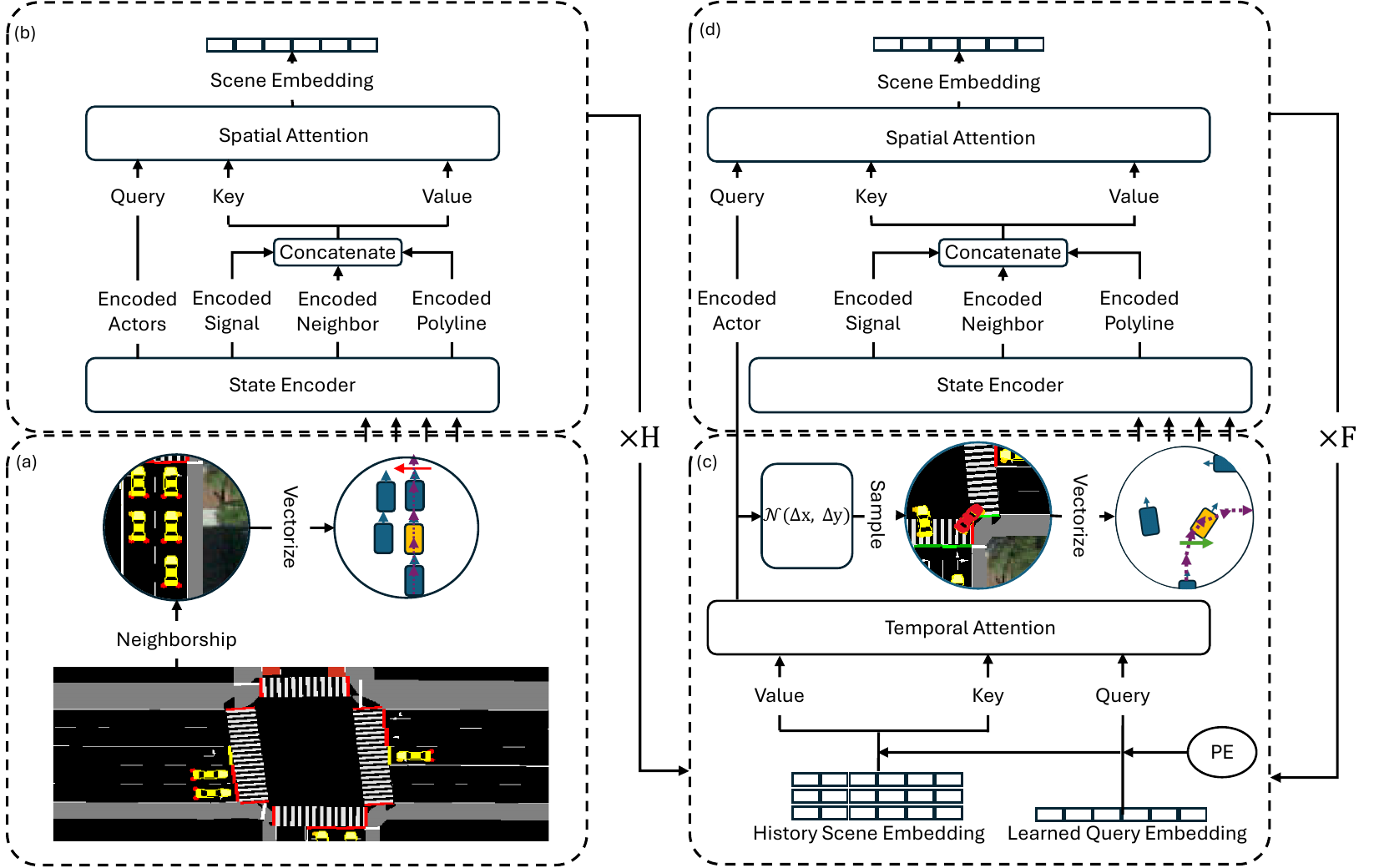}
    \caption{Transformer-based architecture for learning multi-actor interactions and generating physically consistent trajectories. 
    (a) The SUMO simulator initializes the traffic scene based on a set of control parameters. For each vehicle, neighboring actors are identified and their states are vectorized. 
    (b) A state encoder maps the actor and its neighbors into embeddings, which are processed by a spatial attention module to produce interaction-aware spatial embeddings over the previous $H$ timesteps. 
    (c) The sequence of spatial embeddings for each actor is combined with a learned query embedding and passed through a temporal attention module to obtain a temporally informed representation, which is used to parameterize a Gaussian distribution over the actor’s next state. During inference, this distribution is sampled to generate the next state. 
    (d) The newly generated actor state, together with updated neighbor states, is re-encoded and fed back into the model, enabling recursive trajectory unrolling in closed loop.
    }
    \label{fig:example}
\end{figure*}

\section{\uppercase{Architecture}}
When developing an architecture to model driving behavior at intersections, it becomes essential to capture both the spatial and temporal structure of multi-agent interactions. We first use the past $H$ ground-truth states of each actor to identify its neighbors and compute their relative features. For the $F$ future steps, we similarly use ground-truth values during training to maintain consistency in neighborhood selection. At each timestep, we apply self-attention \cite{vaswani2023} to model spatial correlations between an actor and its surrounding agents, map polylines, and signal information. Temporal dependencies are captured through directional attention mechanisms applied across the $F$ predicted steps, producing a Gaussian distribution over each actor’s future state.

\subsection{Spatial Structure}

At any timestep, an actor can observe only a local region of the intersection. To model how spatial interactions influence motion, we encode the spatial structure using self-attention. The query $Q$ is a learned linear projection of the actor’s state, while the keys $K$ and values $V$ are linear projections of its nearest neighbors, nearby map polylines, and signal stop-lines (selected using Euclidean distance):
\begin{align}
    PQ^i_t &= \text{Linear}(o^i_t), \\
    PK^i_t &= \text{Linear}([map^i_t,\, signal^i_t,\, n^{1..N(i,t)}_t]), \\
    PV^i_t &= \text{Linear}([map^i_t,\, signal^i_t,\, n^{1..N(i,t)}_t]), \\
    SE^i_t &= \text{SpatialAttention}(PQ^i_t,\, PK^i_t,\, PV^i_t).
\end{align}

Here, \texttt{SpatialAttention} is a multi-head, multi-layer transformer block that produces a spatial embedding of the actor at each timestep. We compute these embeddings for all $H$ historical timesteps. For each of the $F$ future timesteps, we first compute a temporally informed embedding using directional attention, then apply spatial attention to incorporate interaction effects at that predicted timestep.

\subsection{Temporal Structure}

The future state of an actor depends not only on its own motion history but also on the historical behavior of neighboring agents. To model these temporal dependencies, we apply directional attention over the sequence of spatial embeddings. Following \cite{shi2023mtr}, we use learnable motion queries, combined with cosine temporal embeddings \cite{vaswani2023}, to generate a temporally-aware state embedding from which Gaussian parameters are predicted.

For each future timestep $t$:
\begin{align}
    H^i_t &= SE^i_{1..t-1} + PE_{1..t-1}, \\
    TQ^i_t &= \text{Linear}(M + PE_t), \\
    TK^i_t &= \text{Linear}(H^i_t), \\
    TV^i_t &= \text{Linear}(H^i_t), \\
    PQ^i_t &= \text{TemporalAttention}(TQ^i_t,\, TK^i_t,\, TV^i_t), \\
    SE^i_t &= \text{SpatialAttention}(PQ^i_t,\, PK^i_t,\, PV^i_t),
\end{align}

where $PE$ is the sinusoidal positional encoding and $M$ is a learned motion embedding. The positional query $PQ^i_t$ serves as the final latent representation for timestep $t$, from which we predict the parameters of a Gaussian distribution over $\Delta x$ and $\Delta y$:
\begin{align}
    \mu^{\Delta x}_t,\, \mu^{\Delta y}_t,\, \sigma^{\Delta x}_t,\, \sigma^{\Delta y}_t = \text{Linear}(PQ^i_t).
\end{align}

\subsection{Training and Loss}

The model is trained by minimizing the Negative Log-Likelihood (NLL) of the predicted 2D displacement distribution over a prediction horizon of $F$ future steps. The training loss is computed as the average NLL across all predicted timesteps:
\begin{equation}
    \mathcal{L} = \frac{1}{F} \sum_{k=1}^{F} 
    \mathrm{NLL}\big(P(\Delta x_{t+k}, \Delta y_{t+k}),\, (s_{t+k}, \sin\alpha_{t+k}, \cos\alpha_{t+k})\big),
\end{equation}
where $P(\Delta x, \Delta y)$ denotes the predicted 2D displacement distribution at each future step.

Given the ground-truth speed $s$ and heading angle $\alpha$, the corresponding Cartesian displacements are computed as
\begin{equation}
    \Delta x = s \cos\alpha\Delta t, \qquad
    \Delta y = s \sin\alpha\Delta t.
\end{equation}

We optimize model parameters using the Adam optimizer, combined with a cosine learning rate decay schedule and $4000$ warmup steps, which stabilizes early training and improves long-horizon rollout performance.

\subsection{Data Generation}

We generate training data using the SUMO traffic simulator by constructing digital maps of the target intersections and simulating vehicle motion under realistic traffic conditions. Vehicle trajectories are generated using SUMO’s modified implementation of the Krauss car-following model. Trip definitions are created using the \texttt{randomTrips.py} utility, which samples vehicle routes based on the provided \texttt{net.xml} map and user-specified traffic parameters.

Specifically, we generate vehicle trips with a departure period of $3.0$~s, a fringe factor of $10$, and a binomial distribution parameter of $2$. Traffic signal timing plans are configured to closely match those observed at the corresponding real-world intersections, ensuring that simulated traffic flow remains representative of realistic operating conditions.

To train the proposed model, we simulate $6$ hours ($21{,}600$~s) of traffic data at a temporal resolution of $0.1$~s. This results in a total of $1{,}241{,}600$ training exemplars, computed as the product of the number of vehicles present at each timestep and the total number of simulated timesteps.

\begin{figure*}[t]
    \centering
    \subfloat[]{
        \includegraphics[width=0.48\textwidth]{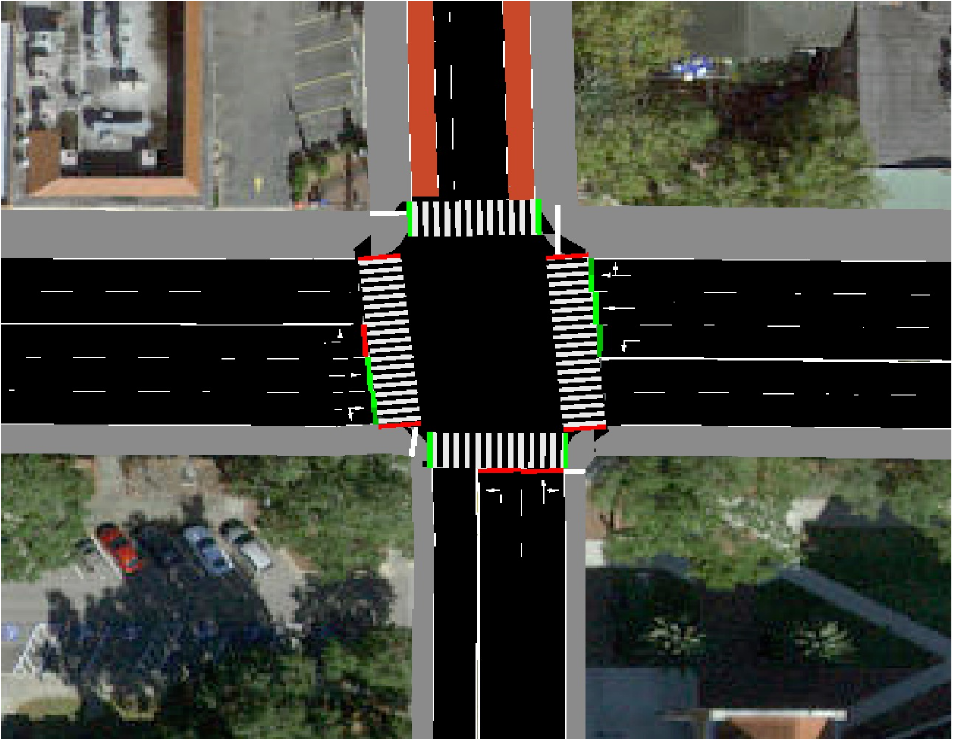}
        \label{fig:img1}
    }
    \hfill
    \subfloat[]{
        \includegraphics[width=0.48\textwidth]{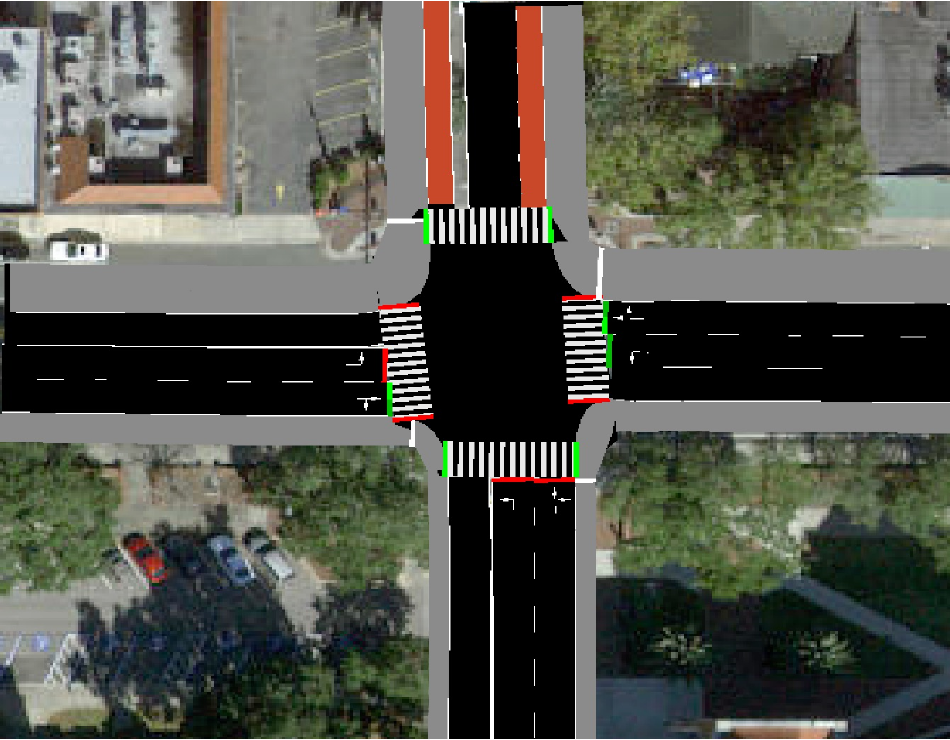}
        \label{fig:img2}
    }
    \caption{The digital representation of two traffic intersection. (a) is the SUMO representation of the intersection at West Univ Ave @ NW 17th Street in Gainesville, Florida. (b) is the same intersection with the middle lane of the east and west flowing traffic removed.}
    \label{fig:two_col}
\end{figure*}

\section{Experiments and Results}

We evaluate the proposed model across two signalized intersections using a simulation-in-the-loop framework. Given an initial intersection configuration and the states of all actors, the learned generative simulator takes over trajectory unrolling in closed loop. These experiments demonstrate the model’s ability to learn the driving behavior across different intersection geometries. Performance is assessed using both intersection-aware safety metrics, which capture rule compliance and interaction quality, and aggregate traffic metrics that summarize macroscopic traffic behavior.

\subsection{Metrics}

\subsubsection{Trajectory Reconstruction Metrics}

Trajectory prediction models are usually evaluated on reconstruction metrics such as:

\begin{itemize}
    \item \textbf{Average Displacement Error (ADE):} The mean $\ell_2$ distance between predicted and ground-truth positions, averaged over all timesteps and all agents.
    \item \textbf{Final Displacement Error (FDE):} The $\ell_2$ distance between the predicted and ground-truth positions at the final prediction timestep $T$.
    \item \textbf{Negative Log Likelihood (NLL):} The negative log likelihood of the ground-truth trajectory under the model's predicted distribution.
\end{itemize}

\subsubsection{Intersection-Aware Metrics}

Prior multi-agent trajectory prediction work has largely focused on offline reconstruction accuracy. However, such metrics fail to capture rule compliance or geometric feasibility of generated trajectories. To address this, we adopt the intersection-aware evaluation metrics proposed in \cite{metrics}, which are aligned with traffic engineering and safety considerations.

\begin{itemize}
    \item \textbf{Red Light Violation}: Counts instances where a vehicle crosses the stop line during a red signal phase.
    
    \item \textbf{Mid-Intersection Stoppage}: Counts occurrences of vehicles stopping inside the intersection box and remaining stationary for a minimum duration.
    
    \item \textbf{Pre-Stopbar Stoppage}: Counts cases where a vehicle fails to cross the stop line during a green phase despite having no upstream blockage.
    
    \item \textbf{Time-To-Collision (TTC) Encounters}: Counts events where the TTC between two vehicles, as reported by the SUMO simulator, falls below a safety threshold.
    
    \item \textbf{Unsafe Deceleration}: Following \cite{vehits23}, we count deceleration events exceeding $-0.47g$, which correspond to hard braking. Here, $g = 9.8\,\mathrm{m/s^2}$.
\end{itemize}

\subsubsection{Aggregate Metrics}

We additionally compute aggregate performance metrics that summarize macroscopic traffic behavior:

\begin{itemize}
    \item \textbf{Average Speed}: The distribution of instantaneous vehicle speeds over the entire simulation horizon. We fit a Gaussian to this distribution and report its mean and standard deviation.
    
    \item \textbf{Average Travel Time}: The mean and variance of travel times for all vehicles that traverse the intersection during the simulation.
\end{itemize}

\begin{table}[t]
    \centering
    \scriptsize
    \setlength{\tabcolsep}{3pt}      
    \renewcommand{\arraystretch}{1.2}

    \begin{tabularx}{\columnwidth}{l *{5}{>{\centering\arraybackslash}X}}
        \rowcolor{black!30}
        \textbf{Model} &
        \textbf{Red Light Violations} &
        \textbf{Mid Intersection Stoppage} &
        \textbf{Pre Stopbar Stoppage} &
        \textbf{Unsafe Acceleration} &
        \textbf{TTC Event} \\
        \rowcolor{gray!15}
        Raw Data      & 0   & 44  & 0 & 0  & 32  \\
        \rowcolor{gray!05}
        IntTrajSim          & 169 & \textbf{1}   & \textbf{0} & 11 & 378 \\
        \rowcolor{gray!15}
        Enactor & 199   & 7  & 37 & 6  & 577 \\
        \rowcolor{gray!15}
        Enactor(N) & \textbf{5}   & 16  & 4 &\textbf{6}  & \textbf{366} \\
    \end{tabularx}

    \caption{Traffic Intersection related Metrics at First Intersection.}
    \label{tab:model_safety_events_1}
\end{table}

\begin{table}[t]
\centering
\scriptsize
\setlength{\tabcolsep}{3pt}
\renewcommand{\arraystretch}{1.15}

\begin{tabularx}{\columnwidth}{l *{5}{>{\centering\arraybackslash}X}}
    \rowcolor{black!30}
    \textbf{Model} &
    \textbf{Mean (m/s)} &
    \textbf{Stand Dev (m/s)} &
    \textbf{KL Divergence} \\
    
    \rowcolor{gray!15}
    Ground Truth & 2.82 & 3.77 & -- \\
    
    \rowcolor{gray!05}
    IntTrajSim & 3.28 & 3.54 & 0.0112 \\

    \rowcolor{gray!15}
    Enactor & 2.87 & 3.59 & \textbf{0.0024} \\
    
    \rowcolor{gray!15}
    Enactor(N) & 2.68 & 3.49 & 0.0063 \\
\end{tabularx}

\caption{Speed distribution statistics at First Intersection.}
\label{tab:speed_stats}
\end{table}

\begin{table}[t]
\centering
\scriptsize
\setlength{\tabcolsep}{3pt}
\renewcommand{\arraystretch}{1.15}

\begin{tabularx}{\columnwidth}{l *{5}{>{\centering\arraybackslash}X}}
    \rowcolor{black!30}
    \textbf{Model} &
    \textbf{Mean (s)} &
    \textbf{Stan.\ Dev.\ (s)} &
    \textbf{KL Divergence} \\
    
    \rowcolor{gray!15}
    Ground Truth & 28.19 & 16.26 & -- \\
    
    \rowcolor{gray!05}
    IntTrajSim & 23.92 & 17.53 & 0.04043 \\

    \rowcolor{gray!15}
    Enactor & 27.24 & 16.03 & 0.00190  \\
    
    \rowcolor{gray!15}
    Enactor(N) & 28.78 & 16.53 & \textbf{0.00093}\\
\end{tabularx}

\caption{Travel time statistics at First Intersection.}
\label{tab:travel_time_stats}
\end{table}

\begin{table}[t]
    \centering
    \scriptsize
    \setlength{\tabcolsep}{3pt}      
    \renewcommand{\arraystretch}{1.2}

    \begin{tabularx}{\columnwidth}{l *{5}{>{\centering\arraybackslash}X}}
        \rowcolor{black!30}
        \textbf{Model} &
        \textbf{Red Light Violations} &
        \textbf{Mid Intersection Stoppage} &
        \textbf{Pre Stopbar Stoppage} &
        \textbf{Unsafe Acceleration} &
        \textbf{TTC Event} \\
        \rowcolor{gray!15}
        Raw Data      & 0   & 41  & 3 & 0  & 42  \\
        \rowcolor{gray!05}
        Enactor(N) & 1   & 82  & 20 & 4  & 534 \\
    \end{tabularx}

    \caption{Traffic Intersection related Metrics at Second Intersection.}
    \label{tab:model_safety_events_2}
\end{table}

\begin{table}[t]
\centering
\scriptsize
\setlength{\tabcolsep}{3pt}
\renewcommand{\arraystretch}{1.15}

\begin{tabularx}{\columnwidth}{l *{5}{>{\centering\arraybackslash}X}}
    \rowcolor{black!30}
    \textbf{Model} &
    \textbf{Mean (s)} &
    \textbf{Stan.\ Dev.\ (s)} &
    \textbf{KL Divergence} \\
    
    \rowcolor{gray!15}
    Ground Truth & 32.31 & 20.60 & -- \\
    
    \rowcolor{gray!05}
    Enactor(N) & 33.38 & 20.01 & 0.00217 \\

\end{tabularx}

\caption{Travel time statistics at Second Intersection.}
\label{tab:travel_time_stats_2}
\end{table}

\begin{table}[t]
\centering
\scriptsize
\setlength{\tabcolsep}{3pt}
\renewcommand{\arraystretch}{1.15}

\begin{tabularx}{\columnwidth}{l *{5}{>{\centering\arraybackslash}X}}
    \rowcolor{black!30}
    \textbf{Model} &
    \textbf{Mean (m/s)} &
    \textbf{Stand Dev (m/s)} &
    \textbf{KL Divergence} \\
    
    \rowcolor{gray!15}
    Ground Truth & 2.52 & 3.69 & -- \\
    
    \rowcolor{gray!15}
    Enactor(N) & 2.35 & 3.29 & 0.0132 \\
\end{tabularx}

\caption{Speed distribution statistics at Second Intersection.}
\label{tab:speed_stats_2}
\end{table}

\subsection{Model Performance}

We evaluate our approach using a simulation-in-the-loop framework following \cite{ranjan2025}. The SUMO microsimulator is calibrated to generate vehicular traffic at a specified arrival rate while controlling the signal phases. For each vehicle, the initial few timesteps are governed by SUMO's built-in car-following model. After this warm-start period, our generative model takes over and begins unrolling trajectories in closed loop. We let this loop run for $40000$ steps (4000 seconds) and generate trajectories of all the vehicles arriving at the intersection. 

For the first intersection as shown in Fig~\ref{fig:img1}, we compare our method against the baseline proposed in IntTrajSim~\cite{ranjan2025} using both interaction-aware safety metrics and the aggregate traffic metrics described earlier. As shown in Table~\ref{tab:model_safety_events_1}, Enactor underperforms the baseline on several safety-related metrics, particularly TTC. Notably, we do not apply the post-processing step used in \cite{ranjan2025}, which likely contributes to this gap. To confirm that our model still learns meaningful vehicle dynamics, we evaluate it on aggregate metrics such as average speed and average travel time. As shown in Table~\ref{tab:speed_stats} and Table~\ref{tab:travel_time_stats}, our model outperforms the baseline by a significant margin---over an order of magnitude in KL divergence. This indicates that the model learns a more accurate distribution over traffic states, even though it initially struggles on interaction-level traffic metrics.

\subsubsection*{\textbf{Ablation Study}}
During qualitative evaluation of the generated trajectories, we observed that vehicles controlled by our model would begin decelerating appropriately when approaching a red signal or a leading vehicle, but often failed to stop at the correct location, overshooting the stop line or the rear of the preceding vehicle. This suggested that the model lacked explicit information about the position of the neighbor's rear bumper, which is crucial for accurate and physically realistic stopping behavior. Incorporating this additional input made the model more aware of its physical surroundings and provided essential contextual information.

This simple modification substantially improved performance on intersection-related safety metrics, as seen in Table~\ref{tab:model_safety_events_1}, where the enhanced model outperforms the baseline on critical metrics such as red-light violations and TTC. Importantly, both variants of our model perform comparably well on the aggregate metrics, further confirming that the core model successfully learns vehicle dynamics even without explicit knowledge of the neighboring vehicle's rear position.

For the second intersection, shown in Fig.~\ref{fig:img2}, we evaluate the model’s ability to perform to a modified intersection geometry. Specifically, the middle lanes for eastbound and westbound traffic are removed. We train the model on data collected from running simulations on the new intersection. As reported in Tables~\ref{tab:model_safety_events_2}--\ref{tab:speed_stats_2}, the proposed model performs favorably across most evaluation metrics. The primary limitation is observed in the Time-to-Collision (TTC) metric, where performance degrades due to occasional instances in which vehicles fail to decelerate sufficiently before reaching a leading vehicle. Despite this, the model maintains strong performance with respect to red-light violations and aggregate traffic metrics, indicating robust rule compliance and realistic macroscopic traffic behavior.

\section{\uppercase{Conclusions}}

In this paper, we present a transformer-based modeling framework for generating physically constrained yet realistic vehicle trajectories at signalized traffic intersections. We evaluate the proposed approach using both traffic-engineering-oriented safety metrics and aggregate measures of speed and travel time. Our results demonstrate that the use of a polar-coordinate representation, combined with closed-loop training, enables the model to more accurately capture vehicle state-transition dynamics. Furthermore, by training and evaluating the model across two distinct intersections, we show its ability to generalize driving behavior across varying intersection geometries.

At the same time, our results highlight opportunities for further improvement, particularly as evidenced by performance on the Time-to-Collision metric. This suggests that more accurate modeling of close-range interactions and deceleration behavior remains an open challenge. Moreover, the development of a more robust and comprehensive set of evaluation metrics that can faithfully assess the realism and safety of simulated traffic remains an important direction for future research. In future work, we aim to investigate transfer learning strategies that enable driving behaviors learned at one intersection to be adapted to new and unseen intersection configurations. Such capabilities would facilitate the study of diverse traffic scenarios while advancing the development of more reliable and expressive evaluation frameworks.

\section*{\uppercase{Acknowledgements}}

The authors acknowledge the use of generative AI tools, specifically OpenAI’s ChatGPT (https://chat.openai.com/), to assist in drafting and refining portions of the manuscript text. AI was used to improve clarity, grammar, structure, and readability of the writing; all scientific content, interpretation, and final revisions were performed by the authors. This acknowledgment is provided in accordance with the conference guidelines on transparent disclosure of AI-assisted writing.

\bibliographystyle{apalike}
{\small
\bibliography{paper}}

@INPROCEEDINGS{intersectionnet,
  author={Wu, Aotian and Ranjan, Yash and Sengupta, Rahul and Rangarajan, Anand and Ranka, Sanjay},
  booktitle={2024 IEEE Intelligent Vehicles Symposium (IV)}, 
  title={A Data-driven Approach for Probabilistic Traffic Prediction and Simulation at Signalized Intersections}, 
  year={2024},
  volume={},
  number={},
  pages={3092-3099},
  doi={10.1109/IV55156.2024.10588424}}

@inproceedings{metrics,
author = {Ranjan, Yash and Sengupta, Rahul and Rangarajan, Anand and Ranka, Sanjay},
title = {Evaluating Generative Vehicle Trajectory Models for nbsp;Traffic Intersection Dynamics},
year = {2025},
isbn = {978-981-96-8294-2},
publisher = {Springer-Verlag},
address = {Berlin, Heidelberg},
url = {https://doi.org/10.1007/978-981-96-8295-9_19},
doi = {10.1007/978-981-96-8295-9_19},
booktitle = {Data Science: Foundations and Applications: 29th Pacific-Asia Conference on Knowledge Discovery and Data Mining, PAKDD 2025, Sydney, NSW, Australia, June 10-13, 2025, Proceedings, Part VI},
pages = {262–274},
numpages = {13},
location = {Sydney, NSW, Australia}
}

@inproceedings{trafficbots2023,
  title={TrafficBots: Towards World Models for Autonomous Driving Simulation and Motion Prediction},
  author={Zhang, L. and Gao, J. and Li, W. and others},
  booktitle={IEEE Intl. Conf. on Robotics and Automation (ICRA)},
  year={2023},
  doi={10.1109/ICRA48891.2023.10160880}
}

@inproceedings{transworldng2023,
  title={TransWorldNG: Traffic Simulation via Foundation Model},
  author={Zhang, Y. and Chen, X. and Liu, Y. and others},
  booktitle={IEEE Intl. Conf. on Smart Computing (SMARTCOMP)},
  year={2023},
  doi={10.1109/SMARTCOMP58114.2023.00076}
}

@article{beyondsimulation2025,
  title={Beyond Simulation: Benchmarking World Models for Planning and Causality in Autonomous Driving},
  author={Zheng, K. and Gao, J. and Li, W. and others},
  journal={arXiv preprint arXiv:2508.01922},
  year={2025}
}

@article{hdgt2023,
  title={HDGT: Heterogeneous Driving Graph Transformer for Multi-Agent Trajectory Prediction via Scene Encoding},
  author={Jia, X. and Wu, P. and Chen, L. and others},
  journal={IEEE Trans. on Pattern Analysis and Machine Intelligence (TPAMI)},
  year={2023},
  doi={10.1109/TPAMI.2023.3298301}
}

@misc{montali2023,
      title={The Waymo Open Sim Agents Challenge}, 
      author={Nico Montali and John Lambert and Paul Mougin and Alex Kuefler and Nick Rhinehart and Michelle Li and Cole Gulino and Tristan Emrich and Zoey Yang and Shimon Whiteson and Brandyn White and Dragomir Anguelov},
      year={2023},
      eprint={2305.12032},
      archivePrefix={arXiv},
      primaryClass={cs.CV},
      url={https://arxiv.org/abs/2305.12032}, 
}

@misc{zhang2025,
      title={Relative Position Matters: Trajectory Prediction and Planning with Polar Representation}, 
      author={Bozhou Zhang and Nan Song and Bingzhao Gao and Li Zhang},
      year={2025},
      eprint={2508.11492},
      archivePrefix={arXiv},
      primaryClass={cs.RO},
      url={https://arxiv.org/abs/2508.11492}, 
}

@misc{gao2020,
      title={VectorNet: Encoding HD Maps and Agent Dynamics from Vectorized Representation}, 
      author={Jiyang Gao and Chen Sun and Hang Zhao and Yi Shen and Dragomir Anguelov and Congcong Li and Cordelia Schmid},
      year={2020},
      eprint={2005.04259},
      archivePrefix={arXiv},
      primaryClass={cs.CV},
      url={https://arxiv.org/abs/2005.04259}, 
}

@misc{vaswani2023,
      title={Attention Is All You Need}, 
      author={Ashish Vaswani and Noam Shazeer and Niki Parmar and Jakob Uszkoreit and Llion Jones and Aidan N. Gomez and Lukasz Kaiser and Illia Polosukhin},
      year={2023},
      eprint={1706.03762},
      archivePrefix={arXiv},
      primaryClass={cs.CL},
      url={https://arxiv.org/abs/1706.03762}, 
}

@misc{shi2023mtr,
      title={Motion Transformer with Global Intention Localization and Local Movement Refinement}, 
      author={Shaoshuai Shi and Li Jiang and Dengxin Dai and Bernt Schiele},
      year={2023},
      eprint={2209.13508},
      archivePrefix={arXiv},
      primaryClass={cs.CV},
      url={https://arxiv.org/abs/2209.13508}, 
}

@conference{vehits23,
author={Rahul Sengupta and Tania Banerjee and Yashaswi Karnati and Sanjay Ranka and Anand Rangarajan},
title={Using {DSRC} Road-Side Unit Data to Derive Braking Behavior},
booktitle={Proceedings of the 9th International Conference on Vehicle Technology and Intelligent Transport Systems - VEHITS},
year={2023},
pages={420-427},
publisher={SciTePress},
organization={INSTICC},
url="https://dx.doi.org/10.5220/0012025300003479",
isbn={978-989-758-652-1},
issn={2184-495X},
}

@misc{ranjan2025,
      title={IntTrajSim: Trajectory Prediction for Simulating Multi-Vehicle driving at Signalized Intersections}, 
      author={Yash Ranjan and Rahul Sengupta and Anand Rangarajan and Sanjay Ranka},
      year={2025},
      eprint={2506.08957},
      archivePrefix={arXiv},
      primaryClass={cs.AI},
      url={https://arxiv.org/abs/2506.08957}, 
}

@article{NI2020102137,
title = {Limitations of current traffic models and strategies to address them},
journal = {Simulation Modelling Practice and Theory},
volume = {104},
pages = {102137},
year = {2020},
issn = {1569-190X},
doi = {https://doi.org/10.1016/j.simpat.2020.102137},
url = {https://www.sciencedirect.com/science/article/pii/S1569190X20300769},
author = {Daiheng Ni}
}

@article{ha2020,
  doi = {10.5281/ZENODO.1207631},
  
  url = {https://zenodo.org/record/1207631},
  
  author = {Ha, David and Schmidhuber, Jürgen},
  
  title = {World Models},
  
  publisher = {Zenodo},
  
  year = {2018},
  
  copyright = {Creative Commons Attribution 4.0}
}

@misc{guan2024worldmodelsautonomousdriving,
      title={World Models for Autonomous Driving: An Initial Survey}, 
      author={Yanchen Guan and Haicheng Liao and Zhenning Li and Jia Hu and Runze Yuan and Yunjian Li and Guohui Zhang and Chengzhong Xu},
      year={2024},
      eprint={2403.02622},
      archivePrefix={arXiv},
      primaryClass={cs.LG},
      url={https://arxiv.org/abs/2403.02622}, 
}

@inproceedings{ettinger2021large,
  title={Large Scale Interactive Motion Forecasting for Autonomous Driving: The Waymo Open Motion Dataset},
  author={Ettinger, Scott and Timofeev, Aleksei and Krivokon, Maxim and Gao, Amy and Joshi, Aditya and Zhang, Yu and Shlens, Jonathon and Anguelov, Dragomir},
  booktitle={Proceedings of the IEEE/CVF International Conference on Computer Vision (ICCV)},
  pages={4378-4387},
  year={2021}
}

@misc{wang2023buildingtransportationfoundationmodel,
      title={Building Transportation Foundation Model via Generative Graph Transformer}, 
      author={Xuhong Wang and Ding Wang and Liang Chen and Yilun Lin},
      year={2023},
      eprint={2305.14826},
      archivePrefix={arXiv},
      primaryClass={cs.LG},
      url={https://arxiv.org/abs/2305.14826}, 
}

@misc{tan2025scenediffusercityscaletrafficsimulation,
      title={SceneDiffuser++: City-Scale Traffic Simulation via a Generative World Model}, 
      author={Shuhan Tan and John Lambert and Hong Jeon and Sakshum Kulshrestha and Yijing Bai and Jing Luo and Dragomir Anguelov and Mingxing Tan and Chiyu Max Jiang},
      year={2025},
      eprint={2506.21976},
      archivePrefix={arXiv},
      primaryClass={cs.LG},
      url={https://arxiv.org/abs/2506.21976}, 
}

@ARTICLE{trajsurvey1,
  author={Huang, Yanjun and Du, Jiatong and Yang, Ziru and Zhou, Zewei and Zhang, Lin and Chen, Hong},
  journal={IEEE Transactions on Intelligent Vehicles}, 
  title={A Survey on Trajectory-Prediction Methods for Autonomous Driving}, 
  year={2022},
  volume={7},
  number={3},
  pages={652-674},
  doi={10.1109/TIV.2022.3167103}}

@misc{salzmann2021,
      title={Trajectron++: Dynamically-Feasible Trajectory Forecasting With Heterogeneous Data}, 
      author={Tim Salzmann and Boris Ivanovic and Punarjay Chakravarty and Marco Pavone},
      year={2021},
      eprint={2001.03093},
      archivePrefix={arXiv},
      primaryClass={cs.RO},
      url={https://arxiv.org/abs/2001.03093}, 
}

\end{document}